# Word Similarity Datasets for Thai: Construction and Evaluation

**PONRUDEE NETISOPAKUL[1], GERHARD WOHLGENANNT[2], AND ALEKSEI PULICH[3]**
[1]Faculty of Information Technology, King Mongkut's Institute of Technology Ladkrabang (KMITL), Bangkok, Thailand (e-mail: ponrudee@it.kmitl.ac.th)
[2]Faculty of Software Engineering and Computer Systems, ITMO University, St. Petersburg, Russia (e-mail: gwohlg@corp.ifmo.ru)
[3]Faculty of Software Engineering and Computer Systems, ITMO University, St. Petersburg, Russia (e-mail: alexpulich@gmail.com)

Corresponding author: Ponrudee Netisopakul (e-mail: ponrudee@it.kmitl.ac.th).

This work is funded by the Academic Melting Pot Program from King Mongkut's Institute of Technology Ladkrabang (KMITL) for the fiscal year 2019. Also, this work was supported by the Government of the Russian Federation (Grant 074-U01) through the ITMO Fellowship and Professorship Program.

**ABSTRACT**

Distributional semantics in the form of word embeddings are an essential ingredient to many modern natural language processing systems. The quantification of semantic similarity between words can be used to evaluate the ability of a system to perform semantic interpretation. To this end, a number of word similarity datasets have been created for the English language over the last decades. For Thai language few such resources are available. In this work, we create three Thai word similarity datasets by translating and re-rating the popular WordSim-353, SimLex-999 and SemEval-2017-Task-2 datasets. The three datasets contain 1852 word pairs in total and have different characteristics in terms of difficulty, domain coverage, and notion of similarity (relatedness vs. similarity). These features help to gain a broader picture of the properties of an evaluated word embedding model. We include baseline evaluations with existing Thai embedding models, and identify the high ratio of out-of-vocabulary words as one of the biggest challenges. All datasets, evaluation results, and a tool for easy evaluation of new Thai embedding models are available to the NLP community online.

## I. INTRODUCTION

The capacity to quantify the degree of semantic similarity between terms is an archetypal way to evaluate the ability of a system to perform semantic interpretation [1]. This operation of lightweight semantic interpretation is applicable in many scenarios, for example to address semantic and lexical gaps in question answering, or in information retrieval ranking operations. Furthermore, semantic similarity between words has applications in many areas such as text summarization, ontology alignment, and machine translation [2]. Word similarity is generally accepted as the most direct intrinsic evaluation metric for word representations. With recent advancements of using neural networks to train low-dimensional semantic representations from large text corpora (coined *word embeddings*) [3], [4], the research field of word representations received heavy attention [2]. Improved word representations provide benefit to most NLP applications that deal with semantics [5].

Some of the well-known existing word similarity datasets include RG-65 [6] (containing 65 word pairs), WordSim-353 [7] (353 pairs), and SimLex-999 [8] (with 999 word pairs). Typically, word similarity gold-standards were ini-tially created for the English language only, although in recent years there have been increased efforts to translate some of the datasets into various other European languages [1], [2], Chinese [9], Indian languages [10], etc. However, for Thai language, to the best of our knowledge, there only exists a very small dataset (65 word pairs) by Osathanunkul et al. [11] based on the Rubenstein & Goodenough's RG-65 [6] dataset – which is too limited in size and other aspects like domain coverage to allow a comprehensive evaluation of Thai word embedding models.

Given the significant role of word similarity datasets in lexical semantics, and the importance of moving beyond the barriers of English language [2], the overall research goal is to provide word similarity datasets for Thai language which are not only high in quality, but also large enough for estimating model performance. Furthermore, the datasets should cover different aspects like the distinction of relatedness and similarity, and involve word similarity tasks of varying difficulty and task characteristics.

The peculiarities of Thai written language, such as missing word and sentence boundaries and flexible word order, make it challenging for NLP. The alphabet includes 44 consonants and 15 basic vowels [11]. Compound vowels can be constructed in various ways by combining vowel characters and consonants, and be placed above, below, before or after the consonants. Further complications in Thai NLP include zero anaphora, the absence of upper/lower-case characters, high ambiguity of compound words, and serial verbs [12]. In contrast to English, Thai is a tonal language with five different tones, making it a very difficult language for English native speakers to comprehend. For this project, the inherent n-gram structure of Thai language had the largest impact on dataset construction and evaluation results. For example, the word *forest*, which is a unigram in English, is translated to *forest-wood* in Thai; *theater* is *hall-drama* in Thai; or *minority* is *race-group-few* in Thai. This n-gram characteristic, coupled with missing word boundaries, makes word segmentation a critical step in the preprocessing pipeline.

With regards to the research goals, we decided to translate the WordSim-353 [7] and SimLex-999 [8] datasets, and the dataset introduced in SemEval-2017 (Task2)[1], subsequently called *SemEval-500* [2]. The translations into Thai language were conducted by two translators for each dataset. In case of disagreement on the translation of specific words a third expert decided. The similarity between terms was rated by 16 (SimLex-999, SemEval-500), and 10 (WordSim-353), resp., native speakers – using the original annotation instructions from the English-language datasets. The final datasets include the mean annotator ratings. We also provide the inter-annotator agreement (IAA) as a human-level baseline, and other statistical data about the datasets. This results in three novel word similarity datasets for Thai language, with 1852 word pairs in total, and different characteristic (regarding task difficulty, domain, relatedness versus similarity) which are inherited from the English-language original datasets. We evaluate existing pre-trained embedding models with the new datasets in order to provide baseline task scores for Thai word embeddings. One finding of the evaluations is the large ratio of out-of-vocabulary (OOV) words – depending on the embedding model. By default, the evaluation tool represents such words by the average vector of the whole vocabulary. We also implemented another strategy, namely piping OOV words into a Thai tokenization tool, and then representing the word as the sum of in-vocabulary parts (if any). The experiments show good Pearson/Spearman correlation scores for in-vocabulary words, so besides improving the quality of word representations themselves, the issue of OOV words is important for future work.

The main contributions of this work include (i) the development of linguistic resources for a low-resourced non-English language, (ii) the translation and rating of three datasets for Thai based on English WordSim-353, SimLex-999 and SemEval-500, (iii) the provision of the datasets including accompanying data such as the fine-grained annotator data and IAA computations, (iv) an evaluation tool which makes it very easy to evaluate any Thai word embedding model with the new datasets, (v) extensive baseline evaluations and a basic variant of dealing with OOV words, and finally, (vi) the analysis and discussion of the specifics of Thai language, esp. with regards to OOV words. The datasets (and accompanying data) are available at: `https://github.com/gwohlgen/thai_word_similarity`, and the evaluation tool for easy evaluation of Thai embeddings can be found at: `https://github.com/gwohlgen/word-embeddings-benchmarks`.

The outline of this publication is as follows: In Section II we discuss existing work related to word similarity datasets and their translation. Section III explains the individual datasets, and the dataset construction process (translation and rating), as well as aspects like inter-annotator agreement. The new datasets are then used to evaluate pre-trained Thai word embedding models in Section IV, in order to provide baseline evaluations and to discuss specifics and issues encountered with Thai language. Finally, we conclude with Section V.

---

[1]alt.qcri.org/semeval2017/task2

## II. RELATED WORK

Earlier work on word representations often used vector-space models of term collocation counts [5], sometimes with postprocessing like dimensionality reduction techniques or the application of term re-weighting (e.g. with PPMI). In contrast to classical one-hot encodings and count-based models, prediction-based models trained with neural networks became very popular since the introduction of word2vec model family [3]. Such word embedding models represent words with low-dimensional, dense, floating-point vectors and are typically trained on large text corpora. Word embedding models have become a crucial ingredient to many NLP systems in the last few years. Other popular algorithms like GloVe and fastText emerged soon after [4], [13]. Word embedding models can either be used standalone, for example as features for word similarity or analogy tasks [14], even in specialized domains like Digital Humanities [15], or they are also commonly used as input layer in ML/deep learning NLP architectures. For the evaluation of model quality, there are two strategies: (i) intrinsic evaluation, where for example human-annotated semantic relations between word pairs are compared to model-predicted relations between these word pairs. The model predictions are generated with similarity functions between term vectors, usually with cosine similarity (see eg. [16], [17]). (ii) The second strategy is extrinsic evaluation of performance on downstream NLP tasks when using specific embedding models as input features [4], [18].

Word similarity datasets are among the most popular intrinsic methods to evaluate distributional semantics models, such as word embedding models [14], [17]. Early datasets like RG-65 [6] (created in 1965) show the long history of this task type. More recent datasets include WordSim-353 [7], SimLex-999 [8], MEN [19], or the datasets introduced in SemEval 2017 (task 2) [2]. Those datasets are basically composed of a list of word pairs and a similarity score for the pair. RG-65 contains 65 word pairs, WordSim-353 contains 353 pairs, and SimLex-999 contains 999 pairs. One main theoretical consideration with word similarity datasets is the distinction between *similarity* and *relatedness*. Similarity refers to synonymy of words or similarity in a number of aspects, whereas relatedness refers to general association. For example "coffee" and "mug" have low similarity, but high relatedness. Some datasets like SimLex-999 provide similarity (synonymy) scores, while others like WordSim-353 provide relatedness scores. Word embedding models are generally better at capturing relatedness than similarity. Some new datasets like SemEval-500 (from SemEval 2017) try to integrate both notions into their scoring via the annotator instructions. For rating the similarity values of word pairs, the typical approach is to ask a number of annotators (usually at least 5 per word pair) for a similarity judgment, and then aggregate the results. This also allows for computing annotator agreement statistics, which are important to compare datasets, and to have a human level baseline for automated methods. In our work, we select WordSim-353, SimLex-999 and SemEval-500 for translation to Thai and subsequent re-rating. WordSim-353 is well established, but for English language automated methods already reach human level agreement scores. SimLex-999 and SemEval-500 are more recent and challenging datasets.

There have been attempts to translate word similarity datasets from their English original to other languages. For Thai there exists little work, except for one project which translated the RG-65 dataset [11] from English to Thai (named TWS-65). For this small dataset, they used a large number of raters (40 Thai native speakers) to assign similarity scores. Osathanunkul et al. evaluate the gold standard dataset against a method based on the structure of WordNet [20]. Regarding other languages, Camacho-Collados et al. [2] published the SemEval-500 dataset in five languages, and Chen and Ma [9] translated SimLex-999 to Chinese. Akhtar et al. [10] translated RG-65 and WordSim-353 into six Indian languages. There have been similar efforts for various languages like Croatian [21], Japanese [22], or Russian [23]. Many researchers used manual rating of the word pairs, while others ( [8], [19], [22], [24]) apply crowdsourcing to translate or rate the datasets with the help of paid crowd workers via crowdsourcing platforms. For the translation of the terms itself, the most popular strategy is to employ two translators for the words, which use high-quality dictionaries to support the translation. Furthermore, the English language similarity scores can help to understand the meaning of polysemous words in the source language. In cases of disagreement between the two translators, often another expert decides.

## III. CONSTRUCTION OF THAI LANGUAGE DATASETS

In this section, we introduce the word similarity datasets, the translation process, as well as the rating of word pairs for the Thai language and inter-annotator agreement.

### A. WORD SIMILARITY DATASETS

As noted by Hill et al. [8], many word similarity datasets like MEN or WordSim-353 give a high score to word pairs that are related by topic or domain. For example *coffee* and *cup* have a high human-assigned similarity rating, although they are only related, but not similar. Motivated by this observation, the authors present the SimLex-999 dataset, which tries to capture similarity. Furthermore, Hill et al. [8] list three criteria which a gold standard dataset for word similarity should satisfy: Firstly, it should be representative, ie. cover the full range of concepts of natural language. Then, it should be clearly defined what the gold standard measures, and finally, native speakers should be able to consistently quantify the target property. The last criteria is typically indicated with inter-annotator agreement.

It is important to note that the rating process for the datasets presented here is context-free, the word pairs are presented to the annotators as-is, without a phrasal or sentential context. Regarding the rating scale for word pairs, most dataset use an absolute scale and specify word similarity for example in the interval of 0 to 10. Some datasets, like MEN [19] are an exception by using relative ranking of word pairs against each other instead.

When using word similarity datasets for evaluating word vectors, one should also be aware of potential limitations, which are analyzed in Faruqui et al. [25]. Firstly, they mention the subjectivity of word similarity judgments, and that human annotators tend to overemphasize relatedness as compared to similarity. Newer datasets like SimLex-999 and SemEval-500 try to mitigate this potential issue. Furthermore, word similarity datasets are usually not split into train and test subsets, which may result in overfitting if models are optimized to solve only the word similarity tasks of specific datasets. Depending on the characteristics of the downstream extrinsic task, performance on word similarity itself may or may not be of crucial. Finally, word similarity datasets, as well as most popular word embedding algorithms, do not address the problem of polysemy of words.

### B. ENGLISH-LANGUAGE DATASETS

A considerable number of English-language word similarity datasets has been released in the last decades, which differ in size, difficulty, rating scale, and other features.

#### 1) RG-65

RG-65 is a classic word similarity dataset presented by Rubenstein and Goodenough [6] already in 1965. It contains only 65 word pairs, and focuses on similarity rather than relatedness. 15 annotators rated the similarity of each word pair.

#### 2) WordSim-353

WordSim-353 [7] is the most popular word similarity dataset [1]. It contains 353 word pairs, and measures semantic *relatedness* on a scale from 0 to 10. WordSim-353 includes two subsets, one set with 153 word pairs rated by 13 annotators, and the remaining pairs rated by 16 annotators. The dataset was later split into relatedness and similarity parts [26]. For English language, state-of-the-art systems have already surpassed human inter-annotator agreement (IAA) for WordSim-353 and RG-65, which can make them unreliable benchmarks for new approaches [2]. For most other languages, esp. languages like Thai, which are still hard for NLP, this limitation is not yet relevant.

#### 3) SimLex-999

As already noted, SimLex-999 [8] is specifically designed to capture similarity between terms. The dataset contains 666 noun, 222 verb and 111 adjective pairs. Characteristics of this dataset are that it includes only words from the vocabulary of WordNet [20], and that it contains a large number of antonymy pairs. The similarity ratings were created with crowdsourcing via Amazon Mechanical Turk, originally on a scale from 0 to 6, later linearly re-scaled to $[0, 10]$.

#### 4) SemEval-500

Camacho-Collados et al. [2] present a multilingual dataset for English, Farsi, German, Italian and Spanish for SemEval-2017, task 2. The dataset contains 500 word pairs. The goal is to provide a challenging dataset, which includes word pairs from 34 domains such as *chemistry and mineralogy, computing*, or *culture and society*. Furthermore, the dataset contains named entities (e.g., Microsoft), and multiword expressions (e.g., black hole) in any of the 34 domains. For rating they use a 5-point Likert scale with a step size of 0.25, and the instructions for the annotators contain both the notions of relatedness and similarity, in which similarity is rated higher.

### C. DATASET CREATION

Three steps were involved in dataset creation: selection of datasets to translate, translation to Thai, and rating of the pairs in the target language. Selecting the datasets, the goal was to choose well-known and popular benchmarks, and to cover various levels of difficulty, and both relatedness and similarity-based datasets. While WordSim-353 contains mostly common terms and concepts and measures relatedness, SimLex-999 focuses on similarity, and finally, SemEval-500 tries to include both the notions of relatedness and similarity, and it includes difficult and rare terms from technical domains as well as named entities and n-grams. Translation and rating is discussed in more details below.

#### 1) Dataset Translation

For the translation from English into other languages, many researchers [2], [9], [10] use the strategy of involving two independent translators, and in case of disagreement on the translations, to have a third expert decide – we applied this approach, too, with translators being Thai academics who are fluent in English. As in Camacho-Collados et al. [2], during translation, the annotators were presented the original similarity score of the word pair, in order to help selecting the correct translation for the intended meanings of the words. Translation agreement between the two initial translators per dataset was low, for example on the WordSim-353 dataset, only for 18.5% of pairs the translators produced the exact same translation of the two words. A number of factors lead to the low agreement. Firstly, polysemy in the source language. One example for this in Table 1 is the English word *stock*, which may refer to company shares, inventory, etc. in English, but the word *stock* only refer to inventory stock for Thais. In this case, the similarity score in the source language did not help the translators. Secondly, Thai language has specific constraints and flexibility in word composition. Foreign words can be translated in multiple ways, for example, *CD* was made into *disk+CD* by one translator and into *disk+record+data* by another. In addition, some words have many translations with the exact same meaning, such as the English word *king* can be translated into a number of different Thai words with exactly the same meaning.

**1.** Comparison between English and Thai datasets with examples from WordSim-353

| English | Thai |
|---|---|
| tiger – cat : 7.35 | เสือ – แมว : 7.05 |
| smart – stupid: 4.62 | ฉลาด – นักเรียน : 4.5 |
| stock – market: 8.08 | คลังสินค้า – ตลาด : 5.4 |
| stock – phone: 1.62 | คลังสินค้า – โทรศัพท์ : 4.8 |

#### 2) Scoring

For scoring the word pairs, we aimed to keep the scores consistent with the original English datasets. We therefore re-used the scoring instructions of the individual English datasets, and the same rating scales (the scales per dataset are described in the description of the English datasets). For the WordSim-353 dataset, 10 native speakers of Thai assigned the similarity scores, for both the SimLex-999 and SemEval-500 scoring was performed by 16 annotators. The annotators rated all word pairs of a dataset. Finally, the resulting scores are computed as the average of the annotator scores. The annotator scoring data for the three new datasets, including some statistical analysis of the scores is available on GitHub[2]

Table 1 provides a few example word pairs in Thai and English. We can see that for the word pairs *tiger–cat* and *smart-stupid* the similarity assessments are very similar, whereas for the last two pairs there are large differences. The polysemous word *stock* was translated with the meaning of *inventory*, which lead to the differences in similarity scores between the source and target language – and it also demonstrates the necessity of re-ranking in the target language.

#### 3) Annotator Agreement

For some datasets like SimLex-999, WordSim-353, and SemEval-500 inter-annotator agreement (IAA) is reported by measuring the average of all pairwise agreement between individual annotators. Others, like Sakaizawa and Komachi [22] apply the average Spearman's $\rho$ between a single annotator and the average of all others. In Table 2 we present the results for both variants, as *Pairwise*, and *To mean*, respectively. Additionally, we show the correlation values between the Thai and the original English versions of the three translated datasets.

Regarding metrics, many authors use Spearman $\rho$ scores to report IAA, and some also provide Pearson's $\rho$. Camacho-Collados et al. [2] use the harmonic mean of these two correlation values as final score. We follow the approach of the latter and present all three values.

---

[2]https://github.com/gwohlgen/thai_word_similarity

**Table 2.** Inter-annotator agreement statistics for the three datasets. Spearman $\rho$ (S), Pearson $\rho$ (P), and harmonic mean of the two (HM).

| Model | TH-WordSim-353 | | | TH-SimLex-999 | | | TH-SemEval-500 | | |
|---|---|---|---|---|---|---|---|---|---|
| | S | P | HM | S | P | HM | S | P | HM |
| Pairwise | 0.581 | 0.585 | 0.583 | 0.646 | 0.691 | 0.668 | 0.702 | 0.706 | 0.704 |
| To mean | 0.728 | 0.736 | **0.732** | 0.782 | 0.819 | **0.800** | 0.827 | 0.826 | **0.826** |
| TH-EN | 0.748 | 0.744 | 0.746 | 0.711 | 0.706 | 0.709 | 0.865 | 0.865 | 0.865 |

In comparison to the IAA scores of the original datasets, the results are as follows: Finkelstein et al. [7] present a Spearman $\rho = 0.61$ for the WordSim-353 dataset for average pairwise correlations. After translation and re-rating, our Spearman $\rho = 0.58$ is similar. For SimLex-999, a Spearman $\rho = 0.67$ was reported [8], as compared to $\rho = 0.65$ of our dataset. Finally, for the SemEval-500 dataset the Spearman $\rho = 0.70$ for Thai. Camacho-Collados et al. report high correlation values between $0.8$ and $0.9$, depending on the language variant of the dataset, however, they use a two-step process of rating, where raters were asked to reassess word pairs for which the rating was distant from the average. We omitted this second step for reasons of consistency within ratings.

Arguably more relevant for the assessment of automated methods are the scores in row *To mean* of Table 2, where raters are compared to the mean of the other raters, in a similar way as distributional semantics models (see evaluations in the next section) are evaluated against the mean of all raters. Therefore, those correlation scores can be viewed as human-level baselines.

## IV. EVALUATION

Given the newly created word similarity datasets, in this section we provide baseline evaluations for Thai word embedding models, as well as a discussion of results.

### A. EVALUATION SETUP

The evaluation setup describes the pre-trained word embeddings used in the evaluations, the evaluation tool and the evaluation metrics.

#### 1) Pre-trained Embedding Models

We use the newly created datasets to provide baseline evaluations for Thai word embedding models. For this purpose, we use pretrained models, which were found by search engine queries and by asking in Thai NLP groups on social media about stock embedding models for Thai. In the following evaluations, we use these five models:

- fastText: fasttext.cc provides models for 157 languages, including Thai[3]. The models are trained on Common Crawl and Wikipedia corpora using fastText [13], regarding settings they report the usage of the CBOW algorithm, 300 dimensions, a window size of 5 and 10 negatives. The model is large and contains 2M vectors. For Thai word segmentation, the ICU tokenizer[4] is applied.
- thai2vec: This model is trained with word2vec on a Wikipedia corpus, and available online[5] as *v.01*. It contains 51K word vectors, with 300 dimensions. For segmentation, a simple dictionary-based approach was used with the pythainlp library[6].
- ft-wiki: We were unable to trace the exact settings used in training this model, it is linked for example here[7]. The model[8] was trained with the fastText library on a Thai Wikipedia corpus. It contains vectors with 300 dimensions for a vocabulary of ca. 108K entries.
- Kyu-ft and Kyu-w2v: The project[9] contains models both trained with fastText (*Kyu-ft*) and word2vec (*Kyu-w2v*). According to the repository the fastText models were trained using the SkipGram algorithm, the word2vec models use CBOW, and a word window of 5. Both models are rather small, with 30K vectors and 300 dimensions.

#### 2) Evaluation Tool and Metrics

For the evaluation of the word embedding models we apply the same metrics like for the computation of IAA in Section III-C3, ie. Spearman correlation between the gold standard dataset and the model output, Pearson correlation, and as final result the harmonic mean of the two values. We reuse and adapt an existing tool for intrinsic evaluation of word embeddings. The tool named "Word Embedding Benchmarks"[10] aims to drive research on word embeddings by providing easy access to evaluation with a lot of existing benchmark datasets [27]. The datasets, however, are currently limited to English language. The similarity of word pairs is computed as the cosine similarity of the corresponding word vectors. More information on installation and usage is found on GitHub.

---

[3]https://fasttext.cc/docs/en/crawl-vectors.html
[4]https://www.elastic.co/guide/en/elasticsearch/plugins/current/analysis-icu-tokenizer.html
[5]https://github.com/cstorm125/thai2fit/
[6]https://github.com/PyThaiNLP/pythainlp
[7]https://github.com/kobkrit/nlp_thai_resources#pre-trained-word-vectors
[8]https://s3-us-west-1.amazonaws.com/fasttext-vectors/wiki.th.vec
[9]https://github.com/Kyubyong/wordvectors
[10]https://github.com/kudkudak/word-embeddings-benchmarks

Table 3. General baseline: All OOV words are replaced by the same average vector. Evaluation metrics Spearman ρ (S), Pearson ρ (P) and harmonic mean (HM) of the two – for the 4 gold standard datasets and all 5 pretrained models. Further, the ratio of OOV words (%OOV) and the number of word pairs with one or two OOV words in it (O-P).

|  | TH-WordSim353 | | | | | TH-SemEval-500 | | | | | TH-SimLex-999 | | | | | TWS65 | | | | |
| --- | --- | --- | --- | --- | --- | --- | --- | --- | --- | --- | --- | --- | --- | --- | --- | --- | --- | --- | --- | --- |
| Model | S | P | HM | %OOV | O-P | S | P | HM | %OOV | O-P | S | P | HM | %OOV | O-P | S | P | HM | %OOV | O-P |
| fastText | 0.182 | 0.179 | 0.181 | 42.1 | 237 | 0.175 | 0.202 | 0.187 | 53.2 | 375 | 0.201 | 0.251 | 0.223 | 35.6 | 550 | 0.203 | 0.147 | 0.170 | 44.6 | 43 |
| thai2vec | 0.384 | 0.331 | **0.356** | 18.4 | 112 | 0.317 | 0.261 | **0.286** | 34.1 | 261 | 0.359 | 0.443 | **0.397** | 7.8 | 146 | 0.505 | 0.504 | **0.505** | 7.7 | 9 |
| ft-wiki | 0.281 | 0.218 | 0.246 | 42.1 | 237 | 0.244 | 0.222 | 0.233 | 53.3 | 375 | 0.292 | 0.287 | 0.289 | 35.6 | 550 | 0.305 | 0.110 | 0.162 | 44.6 | 43 |
| Kyu-ft | 0.331 | 0.208 | 0.256 | 38.5 | 217 | 0.290 | 0.238 | 0.262 | 48.6 | 351 | 0.352 | 0.343 | 0.348 | 31.6 | 502 | 0.526 | 0.410 | 0.461 | 30.8 | 34 |
| Kyu-w2v | 0.252 | 0.193 | 0.219 | 38.5 | 217 | 0.234 | 0.220 | 0.227 | 48.6 | 351 | 0.263 | 0.296 | 0.278 | 31.6 | 502 | 0.497 | 0.481 | 0.489 | 30.8 | 34 |

We forked and adapted the repository specifically for the evaluation of Thai word embeddings. The goal was to make usage very simple – in order to evaluate a new embedding model with the datasets presented here, it is sufficient to add the model file path to the evaluation script, and run it. The repository for evaluating Thai embeddings and reproducing the presented results is available at: https://github.com/gwohlgen/word-embeddings-benchmarks. The original repository only computes Pearson correlation, we made some additions, for example the computation of Spearman ρ, the ability to tokenize terms with deepcut (see below) on demand, or to filter word pairs with OOV words.

As shown in the evaluation results below, we experienced a high number of out-of-vocabulary (OOV) words in the pretrained embeddings, ie. words from the datasets not existing in the embedding vocabulary. As backoff technique, by default, the evaluation tool uses an average vector over all words in the embedding model to represent OOV words.

Sophisticated methods to tackle the problem of OOV words are beyond the scope of this work, but we implemented a baseline method to address the issue using the deepcut tokenizer[11]. Deepcut achieves a F1 score of 98.1% on the BEST dataset for Thai word tokenization. The strategy, which we integrated into the evaluation tool, and which can be applied optionally, is to try to split OOV words into components with deepcut. If any of those components are found in the embedding vocabulary, then the OOV word will be represented by the average vector of the in-vocabulary components.

[11] https://github.com/rkcosmos/deepcut

### B. EVALUATION RESULTS

In this section, we provide the baseline evaluations of the new datasets with the existing pretrained models described in Section IV-A1. We present four result tables, which differ in the way OOV words are handled. Basically all tables show as evaluation metrics the Spearman correlations between the respective gold standard dataset and the embedding model, the Pearson correlation, and as final score the harmonic mean of the two [2].

Table 3 shows the general baseline result for the five pretrained models. Additionally to the correlation scores, the table includes the percentage of OOV words (%OOV), ie. dataset words that do not exist in the vocabulary of the model, and the count of word pairs which contain one or two OOV words (O-P). The evaluation tool used replaces OOV words with an average vector of all words in the vocabulary, which is a vector with no semantic distinctiveness. In Table 3 the correlation metrics are generally low, caused by the high ratio of word pairs with OOV words. For all datasets *thai2vec* yields the best results. This is caused by the lower ratio of OOV words. The best results were achieved for TWS-65 with an $HM = 0.505$. Whereas the three other datasets include mostly common words, and have comparable percentage of OOV words, SemEval-500 contains many technical words and multi-word expressions, which led to a higher fraction of OOV words and the lowest correlation scores.

As already mentioned, we implemented a simple strategy which tries the address the problem of OOV words. Using the deepcut library, OOV words are split into components. If those components are in the vocabulary, the OOV term is represented by the average vector of the components. Table 4 presents the result of this variant. The ratio of OOV words is much lower with this strategy, although some words cannot be split or none of the components are in the vocabulary. But while we measure a percentage of OOV words of up to $53.3\%$ in Table 3, the maximum is now at 22.1%. This simple approach helps to raise the correlation scores to over 0.5 for all datasets. The model *Kyu-ft* consistently provides the best results.

Table 4. Baseline with deepcut tokenizer: OOV words are tokenized with deepcut, and replaced by component vectors, if any. Evaluation metrics as in Table 3

| Model | TH-WordSim-353 | | | | | TH-SemEval-500 | | | | | TH-SimLex-999 | | | | | TWS65 | | | | |
|---|---|---|---|---|---|---|---|---|---|---|---|---|---|---|---|---|---|---|---|---|
| | S | P | HM | %OOV | O-P | S | P | HM | %OOV | O-P | S | P | HM | %OOV | O-P | S | P | HM | %OOV | O-P |
| fastText | 0.347 | 0.363 | 0.355 | 9.2 | 58 | 0.371 | 0.368 | 0.369 | 22.0 | 174 | 0.410 | 0.486 | 0.445 | 10.3 | 188 | 0.252 | 0.200 | 0.223 | 16.9 | 18 |
| thai2vec | 0.471 | 0.433 | 0.451 | 3.3 | 18 | 0.425 | 0.363 | 0.392 | 16.0 | 134 | 0.432 | 0.518 | 0.471 | 1.3 | 25 | 0.530 | 0.589 | 0.558 | 0.0 | 0 |
| ft-wiki | 0.475 | 0.479 | 0.477 | 9.2 | 58 | 0.496 | 0.446 | 0.470 | 22.1 | 175 | 0.505 | 0.551 | 0.527 | 10.3 | 188 | 0.467 | 0.278 | 0.349 | 16.9 | 18 |
| Kyu-ft | 0.572 | 0.527 | **0.549** | 6.7 | 42 | 0.527 | 0.480 | **0.502** | 18.4 | 153 | 0.544 | 0.588 | **0.565** | 8.2 | 148 | 0.754 | 0.718 | **0.735** | 6.9 | 9 |
| Kyu-w2v | 0.456 | 0.477 | 0.466 | 6.7 | 42 | 0.430 | 0.429 | 0.430 | 18.4 | 153 | 0.492 | 0.543 | 0.516 | 8.2 | 148 | 0.686 | 0.687 | 0.687 | 6.9 | 9 |

To better understand the effect of OOV words, in Table 5 we only include word pairs into the evaluations where both terms are in-vocabulary. This greatly reduces the number of evaluation word pairs, the number is shown as *in-vocabulary pairs (IV-P)*. For this subset the correlation scores are very high, for WordSim-353 they even surpass human-level agreement (0.741 vs. 0.732, see Table 2). However, it is dubious that these results hold for the whole dataset in case all OOV words are made in-vocabulary, as the IV-P words tend to be common words. Sahlgren and Lenci [17] show that common words with a high corpus term frequency lead to better vector representations as compared to rarer words.

Finally, we combine the idea of including only in-vocabulary word pairs with deepcut tokenization. The numbers in Table 6 are measured when first trying to replace OOV with their component vectors using deepcut, and then limiting the evaluation to word pairs where both words are in-vocabulary. As expected this increases the number of remaining word pairs substantially (roughly twice as many as in Table 5) but also lowers the correlation metrics.

### C. DISCUSSION OF RESULTS

The evaluations show that the existing Thai word embeddings provide promising results on the word similarity tasks as defined in the new datasets. Overall, amongst the evaluated models, *Kyu-ft* seems to be best suited for NLP tasks which rely on semantic similarity. As expected, the SemEval-500 dataset proved itself to be the most difficult.

The most interesting aspect is the high number of OOV words in the baseline evaluations, which has a large impact on evaluation results. We analyzed and categorized the OOV terms in the datasets. The *thai2vec* embedding consistently provided the lowest number of OOV words, and those words are a subset of the OOV terms of the other models.

The percentages of OOV words for TH-WordSim-353 and TH-SimLex-999 in Table 4 are similar, overall both datasets include similar word categories. Deeper investigation yields various OOV word categories. Firstly, foreign language proper nouns, are often transliterated prefixed with their class, for example *Mexico* is *country+Mexico*. In addition, foreign words, when transliterated into Thai, can be spelled differently using different Thai characters while pronouncing exactly the same sound. Most OOV words from TH-WordSim-353 using thai2vec fall into this category. A second category of OOV words are words with long n-gram translations. For example, *archive* is translated as *place+keep+document+important*, or *OPEC* becomes *group+country+export+oil*. Deepcut is able to segment some of such OOV words into many shorter constituent words. The third category are words which contain a prefix. In Thai, the prefix การ=verb-prefix makes a verb into a noun, and ความ=adjective-prefix provides the same function for adjectives. With the prefix นัก=person-prefix an action or a noun becomes a profession. There is a large number of Thai OOV words in this category, which deepcut can sometimes segment into smaller components. The fourth category of OOV words are technical words, found mostly in the SemEval-500 dataset. Those words are transliterated into Thai with or without a prefix. For example, the word *Joule* is transliterated as is, also the word *Hadoop*.

In summary, although many of the OOV words are very common words in the Thai language, many of those OOV occurrences (esp. from categories 1-3) are caused by mismatches between the dataset words and the output of the Thai word segmentation algorithms. Tools like deepcut tend to cut those dataset terms into multiple components. We will address this problem in future work by training n-gram embedding models (like in Mikolov et al. [3]).

### V. CONCLUSIONS

Semantic word similarity is the most popular task for intrinsic evaluation for models of distributional semantics, such as word embeddings. In this work, we translated three popular

Table 5. Evaluation results for only *in-vocabulary* word pairs. Evaluation metrics as in Table 3, except *IV-P*, which is the count of in-vocabulary pairs.

|  | TH-WordSim-353 | | | | | TH-SemEval-500 | | | | | TH-SimLex-999 | | | | | TWS65 | | | | |
| --- | --- | --- | --- | --- | --- | --- | --- | --- | --- | --- | --- | --- | --- | --- | --- | --- | --- | --- | --- | --- |
| Model | S | P | HM | %OOV | IV-P | S | P | HM | %OOV | IV-P | S | P | HM | %OOV | IV-P | S | P | HM | %OOV | IV-P |
| fastText | 0.750 | 0.732 | **0.741** | 42.1 | 116 | 0.710 | 0.714 | 0.712 | 53.2 | 125 | 0.565 | 0.698 | 0.625 | 35.6 | 449 | 0.817 | 0.783 | 0.800 | 44.6 | 22 |
| thai2vec | 0.554 | 0.520 | 0.537 | 18.4 | 241 | 0.619 | 0.535 | 0.574 | 34.1 | 239 | 0.435 | 0.530 | 0.478 | 7.8 | 853 | 0.541 | 0.594 | 0.566 | 7.7 | 56 |
| ft-wiki | 0.720 | 0.715 | 0.717 | 42.1 | 116 | 0.718 | 0.734 | 0.726 | 53.3 | 125 | 0.591 | 0.711 | 0.645 | 35.6 | 449 | 0.817 | 0.765 | 0.790 | 44.6 | 22 |
| Kyu-ft | 0.749 | 0.710 | 0.729 | 38.5 | 136 | 0.759 | 0.770 | **0.764** | 48.6 | 149 | 0.626 | 0.730 | **0.674** | 31.6 | 497 | 0.860 | 0.846 | **0.853** | 30.8 | 31 |
| Kyu-w2v | 0.719 | 0.698 | 0.708 | 38.5 | 136 | 0.715 | 0.744 | 0.729 | 48.6 | 149 | 0.580 | 0.689 | 0.630 | 31.6 | 497 | 0.758 | 0.743 | 0.750 | 30.8 | 31 |

Table 6. Evaluation results for *in-vocabulary* word pairs after the application of deepcut tokenization. Evaluation metrics as in Table 5.

|  | TH-WordSim-353 | | | | | TH-SemEval-500 | | | | | TH-SimLex-999 | | | | | TWS65 | | | | |
| --- | --- | --- | --- | --- | --- | --- | --- | --- | --- | --- | --- | --- | --- | --- | --- | --- | --- | --- | --- | --- |
| Model | S | P | HM | %OOV | IV-P | S | P | HM | %OOV | IV-P | S | P | HM | %OOV | IV-P | S | P | HM | %OOV | IV-P |
| fastText | 0.526 | 0.527 | 0.526 | 9.2 | 295 | 0.645 | 0.637 | 0.641 | 22.0 | 326 | 0.549 | 0.615 | 0.580 | 10.3 | 811 | 0.748 | 0.708 | 0.728 | 16.9 | 47 |
| thai2vec | 0.524 | 0.507 | 0.515 | 3.3 | 335 | 0.591 | 0.548 | 0.569 | 16.0 | 366 | 0.446 | 0.528 | 0.484 | 1.3 | 974 | 0.530 | 0.589 | 0.558 | 0.0 | 65 |
| ft-wiki | 0.544 | 0.555 | 0.549 | 9.2 | 295 | 0.690 | 0.696 | 0.693 | 22.1 | 325 | 0.584 | 0.636 | 0.609 | 10.3 | 811 | 0.766 | 0.694 | 0.728 | 16.9 | 47 |
| Kyu-ft | 0.622 | 0.591 | **0.606** | 6.7 | 311 | 0.700 | 0.698 | **0.699** | 18.4 | 347 | 0.617 | 0.660 | **0.638** | 8.2 | 851 | 0.842 | 0.780 | **0.809** | 6.9 | 56 |
| Kyu-w2v | 0.574 | 0.584 | 0.579 | 6.7 | 311 | 0.669 | 0.684 | 0.676 | 18.4 | 347 | 0.600 | 0.656 | 0.627 | 8.2 | 851 | 0.810 | 0.786 | 0.798 | 6.9 | 56 |

English language word similarity datasets (WordSim-353, SimLex-999, and SemEval-500) into Thai language. Those datasets are diverse regarding their aim (similarity versus relatedness), difficulty, and term properties. Baseline evaluations with pretrained word embedding models analyse the quality of existing models and we study potential issues, esp. the problem of OOV words.

The main contributions are the development of linguistic resources for the Thai language in form of the datasets TH-WordSim-353, TH-SimLex-999 and TH-SemEval-500 by translating and re-rating the English originals, furthermore providing the datasets and the detailed annotator information to the NLP community, the creation of an evaluation tool which makes it easy to evaluate a given Thai word embedding model with the new datasets, and finally we conducted extensive baseline evaluations with pretrained models and discussed Thai-language specific findings.

Regarding future work, the first angle of approach is improving model quality by training models using the latest NLP stack for Thai, for example the deepcut tokenizer to segment the training corpus. This might also help reduce the ratio of OOV terms. In general the reduction of OOV words is essential, multiple routes are promising here: training n-gram embeddings as eg. in Mikolov et al. [3], use fastText's feature of providing vectors for OOV words from subword features, or try other subword-embeddings such as byte-pair encodings (BPEmb) [28]. In the SemEval 2017 (task 2) evaluations, the top-performing participants combined distributional language models with information from knowledge resources. The amount of knowledge resources is limited for Thai, but for example a Thai WordNet exists. Finally, another interesting aspect of studying Thai word embeddings will be an experimental study on the relation of corpus size and embedding quality, as done for English language by Sahlgren and Lenci [17].

### ACKNOWLEDGMENTS
This work is funded by the Academic Melting Pot Program from King Mongkut's Institute of Technology Ladkrabang (KMITL) for the fiscal year 2019. Also, this work was supported by the Government of the Russian Federation (Grant 074-U01) through the ITMO Fellowship and Professorship Program.

. . .